\documentclass{IEEEoj-data}
\usepackage{cite}
\usepackage{amsmath,amssymb,amsfonts}
\usepackage{algorithmic}
\usepackage{graphicx,color}
\usepackage{textcomp}
\usepackage{hyperref}
\usepackage{balance}

\usepackage{subcaption}
\usepackage{booktabs}
\usepackage{makecell}
\usepackage{tikz}
\usepackage{amssymb}
\usepackage[locale = UK]{siunitx}
\usepackage{colortbl}
\usepackage{graphicx}
\usepackage{forest}
\usepackage{tabularx}
\usepackage{multirow}

\def\BibTeX{{\rm B\kern-.05em{\sc i\kern-.025em b}\kern-.08em
    T\kern-.1667em\lower.7ex\hbox{E}\kern-.125emX}}
\AtBeginDocument{\definecolor{ojcolor}{cmyk}{0.93,0.59,0.15,0.02}}

\begin{document}

\title{\textcolor{black}{Descriptor:} \textcolor{ieeedata}{\textit{LYNRED Mobility Dataset: Multimodal Detection Subset (LYNRED-MDS)}}}

\author{\;
Loïc Arbez\authorrefmark{1} , Jessy Matias\authorrefmark{2}, Xavier Brenière\authorrefmark{2}, Jocelyn Chanussot (FELLOW, IEEE)\authorrefmark{1}, Ronald Phlypo\authorrefmark{3} 
}

\affil{INRIA center at University Grenoble Alpes, 38330 Montbonnot-Saint-Martin, France}
\affil{Lynred, 38113, Veurey-Voroize, France}
\affil{Univ Grenoble Alpes, CNRS, Grenoble INP, GIPSA-lab, 38000 Grenoble, France}

\corresp{CORRESPONDING AUTHOR: Loïc Arbez (e-mail: loic.arbez@inria.fr).}
\authornote{This work was supported by the Fondation Grenoble INP : DeepRed Research Chair, under the patronage of Lynred.}
\markboth{DESCRIPTOR : LYNRED Mobility Dataset : Multimodal Detection Subset }{ARBEZ}

\begin{abstract}

Current road safety systems primarily focus on minimizing post-collision damage. However, advances in algorithmic perception are shifting focus toward early collision prediction, especially in low-visibility conditions like nighttime or fog, where thermal infrared sensing outperforms both human vision and RGB imaging. While available RGB-infrared datasets such as FLIR ADAS and LLVIP are good benchmarks, they mostly consist of clear weather and overly simple scenarios. In this paper, we introduce the LYNRED-MD : Multimodal Detection Subset, a subset of the LYNRED Mobility Dataset, comprised of 4000 RGB-infrared image pairs captured under diverse weather, lighting, and road conditions around Grenoble, France. Our dataset spans varied driving contexts (urban, rural, mountainous, etc.) and a vehicle fleet compliant with Western European standards. Thermal cross-dataset evaluation using a YOLOv8n baseline suggests that our dataset offers strong generalization potential for pedestrian detection in driving scenarios. By covering critical edge cases, our dataset supports the development of more reliable and deployable vision systems for advanced driver-assistance systems.\\

\,\\

 {\textcolor{ieeedata}{\abstractheadfont\bfseries{IEEE SOCIETY/COUNCIL}}} IEEE Intelligent Transportation Systems Society \\  
 \\
 {\textcolor{ieeedata}{\abstractheadfont\bfseries{DATA DOI/PID}}}   https://www.lynred.com/lynred-mobility-dataset  \\ 
  
 {\textcolor{ieeedata}{\abstractheadfont\bfseries{DATA TYPE/LOCATION}}}  Thermal images, Visible images,  Multispectral images; Grenoble, Auvergne-Rhône-Alpes, FRANCE

\end{abstract}


\maketitle

\section*{BACKGROUND} 
Recent advancements in technology have triggered a significant shift in regulatory frameworks designed to enhance the safety of vulnerable road users (VRUs). These frameworks increasingly focus on the integration of advanced technologies, particularly in the realm of vehicle safety systems. With technologies such as computer vision-based pedestrian and cyclist detection systems proving effective in reducing VRU-related collisions~\cite{tomtom, NHTSA, EUGSR}, many jurisdictions now require the inclusion of advanced driver assistance systems (ADAS) in new vehicles. For example, the European Union's General Safety Regulation now requires features like automated emergency braking tailored for VRU scenarios~\cite{EUGSR}, and organizations like the European New Car Assessment Programme have incorporated these systems into their vehicle safety assessment protocols~\cite{tomtom}. In the United States, the National Highway Traffic Safety Administration similarly emphasizes the adoption of ADAS technologies, in particular for Automatic Emergency Braking (AEB) systems~\cite{NHTSA}. In that regard, they require these AEB systems to avoid impact at 62mph (about 100km/h). This means emergency braking systems should be able to detect objects, in this case human beings, at about 50 meters assuming an AEB reaction time close to instantaneous.\newline

\begin{figure*}   
   \centering
   \includegraphics[width=0.32\textwidth]{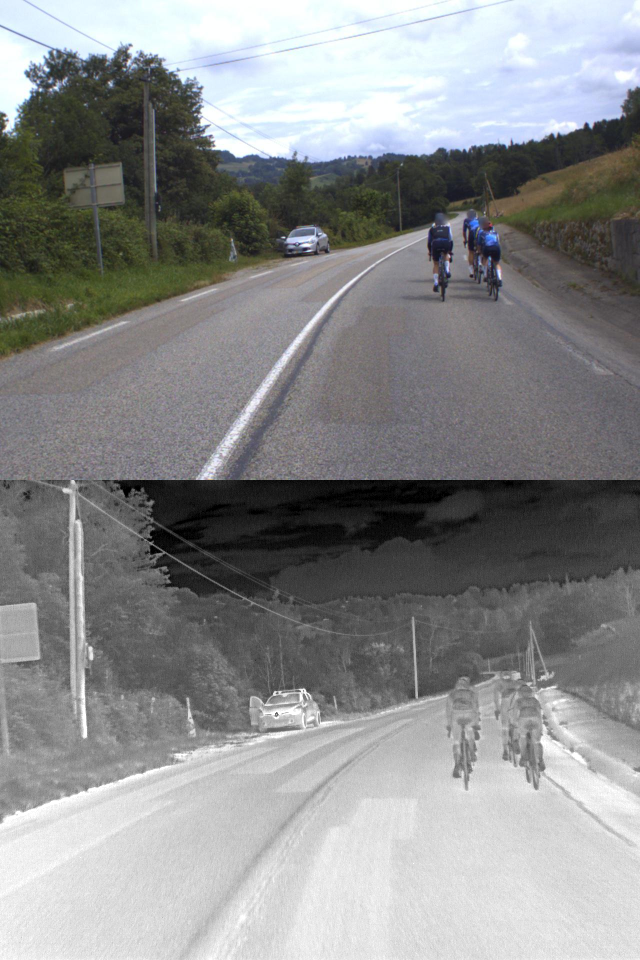}
   \hfill
   \includegraphics[width=0.32\textwidth]{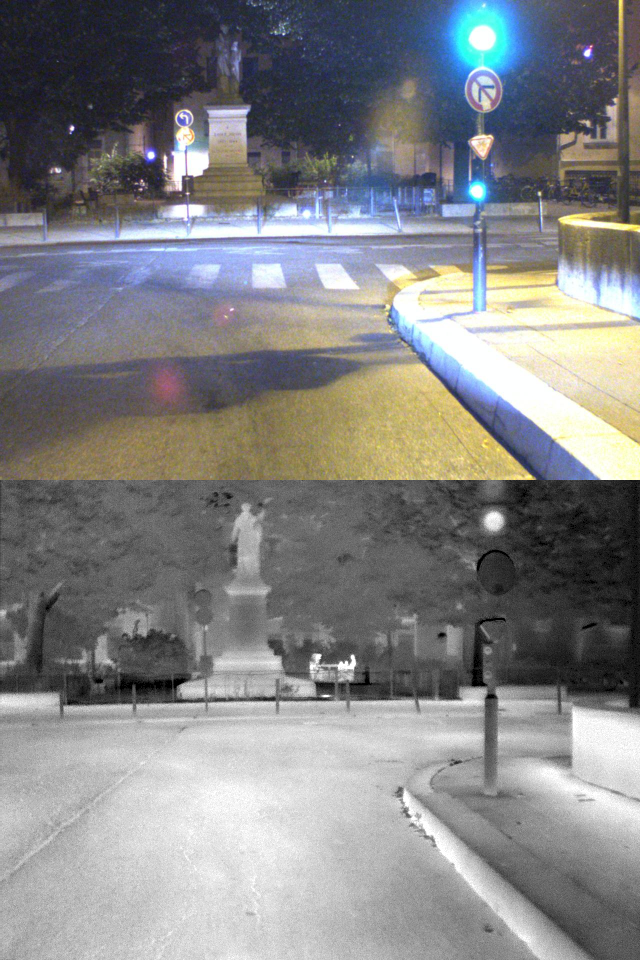}%
   \hfill
   \includegraphics[width=0.32\textwidth]{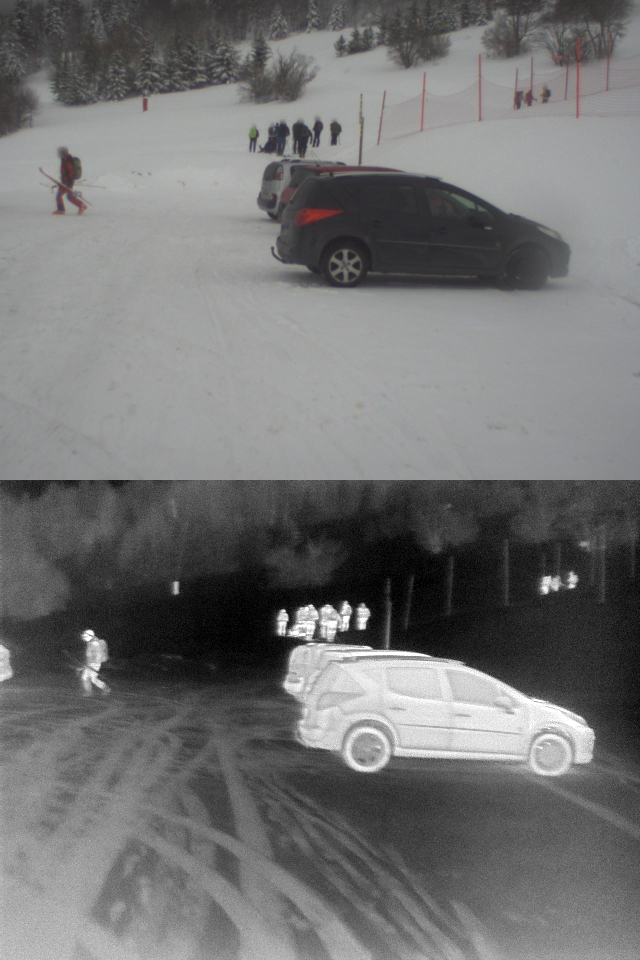}
  \caption{Three samples of the multimodal detection subset of the LYNRED mobility dataset showing the diversity of the dataset.}
  \label{fig:Extracts}
\end{figure*}

In addition to vehicle-based systems, regulatory efforts have extended to urban infrastructure. Many cities have introduced mandates for adaptive traffic controls, smart crosswalks, and reduced speed limits in high--density VRU zones~\cite{EUGSR}. These efforts build on traditional safety measures, such as crashworthiness standards and pedestrian-friendly design and further reflect a broader shift toward leveraging cutting-edge technologies like computer vision and smart sensors to address VRU safety in a more comprehensive manner. This shift toward predictive safety has been facilitated by advancements in computer vision and sensor technologies, which have both reduced the cost of sensors and improved the robustness of detection systems over the last decades. The automotive industry has, since then, embraced these technologies.\\ 

The publication of public datasets sparked further research in the field. Notable among these are the Kitti~\cite{Geiger2012CVPR} and Cityscapes~\cite{Cordts2016Cityscapes} datasets, which introduced demanding tasks such as semantic segmentation and depth estimation, mostly leveraging light detection and ranging (LiDAR) data. Alongside those active sensor technologies, several thermal imaging datasets made their way into the pedestrian or VRU detection community, demonstrating the efficacy of passive IR imaging under various lighting conditions. Indeed, long-wave infrared (LWIR) or thermal sensors---unlike visible light or RGB sensors---do not rely on external light sources such as the sun or a car's headlights, nor on emitted signals like LiDAR or ultrasound sensors. Indeed, infrared imaging uniquely relies on the thermal radiation of objects, i.e. the quantity of photons released by a given body as a function of its temperature~\cite{corsi2015new}. This makes it an ideal candidate for pedestrian detection tasks. As a matter of fact, humans have an internal temperature of 36.6\si{\celsius}~\cite{ley2023defining} and hence tend to be easily distinguishable from their environment~\cite{gonzalez2016pedestrian,dona2024thermal}.\\

The collection of visible data being relatively inexpensive, these IR datasets are often paired with visible RGB sensor data. Most modern vehicles are already equipped with a standard optical camera. As a result, RGB-T (visible-thermal) or RGB-IR (visible-infrared) datasets have been released. The first works on those paired visible and thermal images in object detection were through datasets such as KAIST multi-spectral~\cite{hwang2015multispectral} and CVC-14~\cite{gonzalez2016pedestrian}. The KAIST multispectral dataset has been particularly influential in the development of VRU detection tasks. Following KAIST, the FLIR thermal ADAS dataset~\cite{FLIR} helped further establish the utility of thermal-visible pairs for driving scenarios~\cite{wilson2023recent}. It succeeded in raising awareness for VRU detection and paved the way for this new paradigm~\cite{wilson2023recent}. In addition to the on-vehicle datasets, more recent static datasets like LLVIP~\cite{jia2021llvip} have been published. Although they differ from the original focus on RGB-T object detection for on-vehicle scenarios, they provide valuable insights for research in RGB-T object detection. Additionally, some datasets have introduced new tasks for infrared modalities, such as depth estimation alongside object detection~\cite{franchi2024infraparis}.\\

Despite the extensive literature on thermal datasets, most of the existing works focus on favorable detection scenarios, primarily---if not exclusively---using ideal weather conditions for both thermal and visible modalities. Extreme weather conditions and challenging real-world environments are rarely represented. Yet, the systems developed using these datasets are expected to perform well even under such conditions. To address this gap, we present a recently published RGB-T object detection dataset: the multimodal detection subset of the LYNRED Mobility Dataset (LYNRED-MD). The latter offers images acquired during more diverse and realistic driving scenarios. A few snapshot pairs are presented in Figure~\ref{fig:Extracts}.

\subsection*{Related datasets}
\label{sec:related}

In this section, we present the state of the art datasets, their specifications and assess their contribution to the literature. A summary can be found in Table~\ref{tab:datasetsspecs}.

\subsubsection*{KAIST Multi-spectral}
Published in 2015 by the Korean Advanced Institute of Science and Technology, it includes over 95k aligned color-thermal pairs of images and 103,128 annotations in its first version \cite{hwang2015multispectral}.The original test split annotations were first criticized for their "problematic bounding boxes" and corrected by \cite{Liu2017}, while \cite{li2018multispectral} subsequently extended this effort to the training split. Both corrected splits are now commonly distributed through \cite{li2018multispectral}. Despite its relatively low resolution by today's standards, this cleaned version remains widely used for model evaluation and comparison.

\subsubsection*{FLIR thermal ADAS}

Proposed by Teledyne FLIR, it is an Advanced Driver Assistance System dataset. Its second version is composed of 13k pairs of RGB-T images, offered both in 16 and 8-bit for the infrared images. Initially annotated with 5 classes, it was extended to 16 classes with the release of its second version in 2022~\cite{FLIR}. It presents images acquired mostly in favorable weather conditions, where neither the visible nor the infrared sensors are significantly challenged. The dataset was criticized by~\cite{zhang2020multispectral} for offering unaligned annotations for the RGB and infrared modalities. The misaligned images were eventually removed, leading to the FLIR aligned subset of the original FLIR ADAS dataset.

\subsubsection*{LLVIP}

Although it consists only of static video surveillance images, the LLVIP dataset~\cite{jia2021llvip} is another recent dataset that caught the community's interest thanks to its size and the quality of its images and labels. It makes a good benchmark for multispectral object detection models. Furthermore, the camera setup used offers optically aligned infrared and visible images, hence removing the geometrical alignment constraint that most other datasets of its kind suffer from. This explains the growing attention this dataset has received since its release in 2021.

\begin{table*}[!htbp]
    \centering
    \resizebox{0.99\textwidth}{!}{
        \begin{tabular}{lrcccccccccc}
            \hline

            \multicolumn{3}{c}{} & \multicolumn{2}{c}{Camera setup} & \multicolumn{2}{c}{\makecell{Image resolution \\ (Width $\times$ Height)}} & \multicolumn{2}{c}{\makecell{Intermodal \\
                Alignment}} & \multicolumn{3}{c}{\makecell{Metadata}}   \\ 
            
            \cmidrule(lr){4-5} \cmidrule(lr){6-7} \cmidrule(lr){8-9} \cmidrule(lr){10-12}
            
            Datasets &  \makecell[c]{\# train \\ image pairs } &  \# classes &  On vehicle &  Stationary &  IR &  RGB &  Fully Aligned &  Chessboard &  \makecell{Outside\\temperature} &  Season &  Occlusion \\

            \hline
            
            KAIST sanitized \cite{chen2024amfd} & \num{47.5}k  & 1 & \checkmark & — & 640 $\times$ 480 & 640 $\times$ 480 & \checkmark & — & — & — & \checkmark \\ 
            FLIR ADAS \cite{FLIR} & 13k\phantom{.0} & 16 & \checkmark & — & 640 $\times$ 512 & 1800 $\times$ 160 $^\star$ & — & \checkmark & — & — & \checkmark \\  
            LLVIP \cite{jia2021llvip} & 15k\phantom{.0} & 1 & — & \checkmark & 1080 $\times$ 720 & 1080 $\times$ 720 &\checkmark & — & — & — & —  \\ 
            M3FD \cite{m3fd} & \num{4.2}k\phantom{.0} & 6 & \checkmark & \checkmark & 1024 $\times$ 768 $^\star$ & 1024 $\times$ 768 $^\star$ & \checkmark & — & — & — & — \\ 
            FLIR aligned \cite{zhang2020multispectral} & 5k\phantom{.0} & 3 & \checkmark & — & 640 $\times$ 512 & 640 $\times$ 512 & — & \checkmark & — & — & — \\
            
            
            LYNRED-MDS (ours) & 4k\phantom{.0} & 9 & \checkmark & — & 640 $\times$ 480 & 1280 $\times$ 960 & — & \checkmark & \checkmark & \checkmark & \checkmark \\ \hline
        \end{tabular}
    }
    \caption{Datasets specifications for all the presented RGB-T datasets. We use ($^\star$) to signal that image size is variable. In this case, only the most frequent size is displayed.}
    \label{tab:datasetsspecs}
\end{table*}

\subsubsection*{M3FD}
Proposed by Liu et al.~\cite{m3fd}, it consists of both static and on-vehicle images captured in Dalian, China. This dataset offers 4.2k RGB-T pairs for training and 300 for testing, and 6 annotated classes (People, Car, Bus, Motorcycle, Lamp, Truck) representing most road users. Even though it is on the smaller side of the spectrum for RGB-T datasets, the diversity of scenes offered, as well as the associated annotations on weather and lighting conditions, make it an interesting case study for real-world ADAS scenarios. In addition to the original annotations, \cite{Deevi_2024_WACV} offers further labels for weather conditions that can be useful for evaluation purposes.

\subsubsection*{Honorable mentions}

Among existing datasets, InfraParis~\cite{franchi2024infraparis} is the most similar in scope to ours, as it provides annotations for both infrared and RGB images acquired around Paris, France. However, due to significant differences in the fields of view between the two modalities, it is unsuitable for multispectral or multimodal object detection tasks. Another notable dataset is MFNet~\cite{takumi2017multispectral}, which offers both bounding box annotations and semantic segmentation masks for a variety of road users in urban environments in Tokyo, Japan. Lastly, the SJTU Multispectral Object Detection (SMOD) Dataset~\cite{chen2024amfd} focuses on scenes with a high density of vulnerable road users, particularly pedestrians and cyclists, making it well-suited for safety-critical perception research.

\section*{COLLECTION METHODS AND DESIGN} 
\subsection*{Data collection setup description}

The multimodal detection subset of LYNRED-MDS presented in this paper, comprises 4,000 RGB–infrared image pairs collected from 12 driving sequences recorded around Grenoble, France. Data acquisition was performed using two sensor pairs—each consisting of an infrared and an RGB camera—mounted on an aluminium frame fixed to the roof of a test vehicle. \\
 
Infrared data was acquired using two different infrared camera sensors, each contributing approximately half of the provided images. The first being a LYNRED PICO640Gen2 sensor, used for 2518 images while the remaining 1482 were acquired with a LYNRED ATTO640D-02. The ATTO sensor is equipped with optics providing a $30^\circ$ horizontal field of view and an aperture of $f/1.2$. While the PICO sensor is mounted with a $42^\circ$ $f/1.2$ lens. This setup ensures consistent framing across the different sensors while maintaining sensitivity under low--light and long--range conditions.\\

The RGB cameras integrate a Sony IMX273 sensor (1448$\times$1086 pixels) with a lens allowing a horizontal field of view of $45^\circ$ and an aperture of $f/1.4$. Given the difference in field of view between the RGB and IR camera sensors, RGB images are cropped from 1448$\times$1086 pixels to 1280$\times$960 pixels retaining only the rightmost and vertically centered region as it represents the region covered by their paired IR sensors.\\

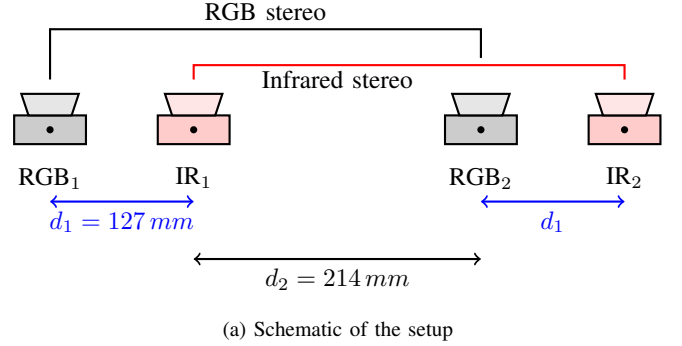
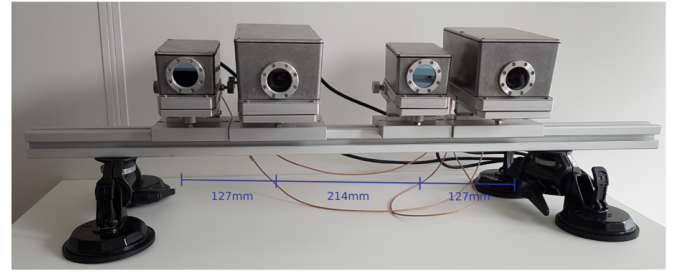
\begin{figure}[!htbp]
    \centering
    \begin{subfigure}{0.49\textwidth}
    
    \begin{tikzpicture}
    \centering
    \begin{scope}[scale=0.95, transform shape]
        
        \draw[thick, fill=black!20] (-4.5, 0) rectangle (-3.5, -0.4); 
        \draw[thick, fill=black!10] (-4.3, 0) -- (-4.4, 0.3) -- (-3.6, 0.3) -- (-3.7, 0) -- cycle; 
        \node[below, yshift=-0.1cm] at (-4, -0.5) {$\text{RGB}_1$};
        \node[fill=black, circle, inner sep=1pt] at (-4, -0.2) {}; 
        
        \draw[thick, fill=red!20] (-2.5, 0) rectangle (-1.5, -0.4); 
        \draw[thick, fill=red!10] (-2.3, 0) -- (-2.4, 0.3) -- (-1.6, 0.3) -- (-1.7, 0) -- cycle; 
        \node[below, yshift=-0.1cm] at (-2, -0.5) {$\text{IR}_1$};
        \node[fill=black, circle, inner sep=1pt] at (-2, -0.2) {}; 
        
        \draw[thick, fill=black!20] (1.5, 0) rectangle (2.5, -0.4); 
        \draw[thick, fill=black!10] (1.7, 0) -- (1.6, 0.3) -- (2.4, 0.3) -- (2.3, 0) -- cycle; 
        \node[below, yshift=-0.1cm] at (2, -0.5) {$\text{RGB}_2$};
        \node[fill=black, circle, inner sep=1pt] at (2, -0.2) {}; 
        
        \draw[thick, fill=red!20] (3.5, 0) rectangle (4.5, -0.4); 
        \draw[thick, fill=red!10] (3.7, 0) -- (3.6, 0.3) -- (4.4, 0.3) -- (4.3, 0) -- cycle; 
        \node[below, yshift=-0.1cm] at (4, -0.5) {$\text{IR}_2$};
        \node[fill=black, circle, inner sep=1pt] at (4, -0.2) {}; 
        
        \draw[<->, thick, blue] (-4, -1.2) -- (-2, -1.2) node[midway, below] {$d_1 = 127\,mm$};
        \draw[<->, thick, blue] (2, -1.2) -- (4, -1.2) node[midway, below] {$d_1$};
        \draw[<->, thick, black] (-2, -2) -- (2, -2) node[midway, below] {$d_2 = 214\,mm$};
        
        \node[above] at (-1, 1.2) {RGB stereo};
        \draw[-, thick, black] (-4, .5) -- (-4, 1.2) -- (2, 1.2) -- (2, 0.8) ;
        \node[below] at (0, .75) {Infrared stereo};
        \draw[-, thick, red] (-2, .5) -- (-2, .7) -- (4, .7) -- (4, 0.5) ;
    \end{scope}
        
    \end{tikzpicture}
    \caption{Schematic of the setup}
    \end{subfigure}
    \hfill

    \begin{subfigure}{0.49\textwidth}
        \includegraphics[width=1\textwidth]{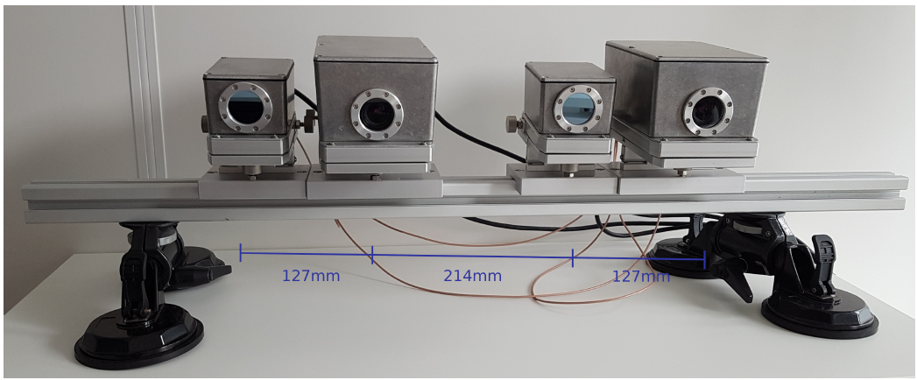}
        \caption{Photograph of the actual setup on the roof of the car}
    \end{subfigure}
    \caption{Schematic representation of the double stereo setup used for the dataset acquisition. For the object detection problem, only the left pair is kept. In order to get an image aligned in the zone of interest, the sensors are angled such that object at infinity (in our case over 10m) are aligned on all sensors.}
    \label{fig:setero_setup}
\end{figure}

The system's geometry is described in Figure~\ref{fig:setero_setup}. The main RGB camera ($\text{RGB}_1$ on Figure~\ref{fig:setero_setup}) controls a trigger signal for the three other cameras to obtain synchronous acquisition from each sensor. Images are acquired at a standard frame rate of 30 images per second, and one image every 3 seconds is kept to maximize diversity in the training set. The sensors were initially oriented such that objects situated 10\,m away from the vehicle and 1\,m above the ground are approximately centered in all sensors’ fields of view. This configuration was chosen to ensure good alignment between modalities for relevant detection targets.\\

\subsection*{Dataset content description}
\label{sec:dataset_description}

This dataset is---to the best of our knowledge---the first large--scale European ADAS dataset specifically focusing on VRU detection in multimodal infrared imagery. LYNRED-MDS consists of a total of 4,000 images, divided into 3,168 image pairs for training and 832 for testing. The data was collected in the surroundings of Grenoble (France). It includes a diverse range of driving scenarios, from winter snowy ski resorts to sunny summer urban areas.\\

\begin{figure*}[tbp]
    \centering
    \begin{subfigure}[t]{0.32\linewidth}
        \centering
        \includegraphics[width=\linewidth]{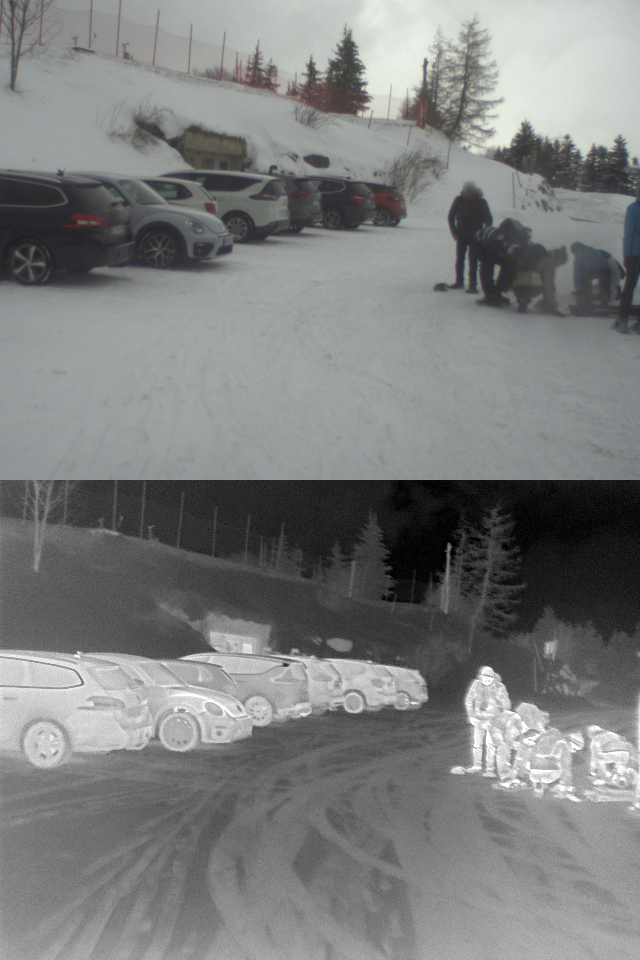}
        \caption{External temperature: \num{5.4}\si{\celsius}}
    \end{subfigure}
    \hfill
    \begin{subfigure}[t]{0.32\linewidth}
        \centering
        \includegraphics[width=\linewidth]{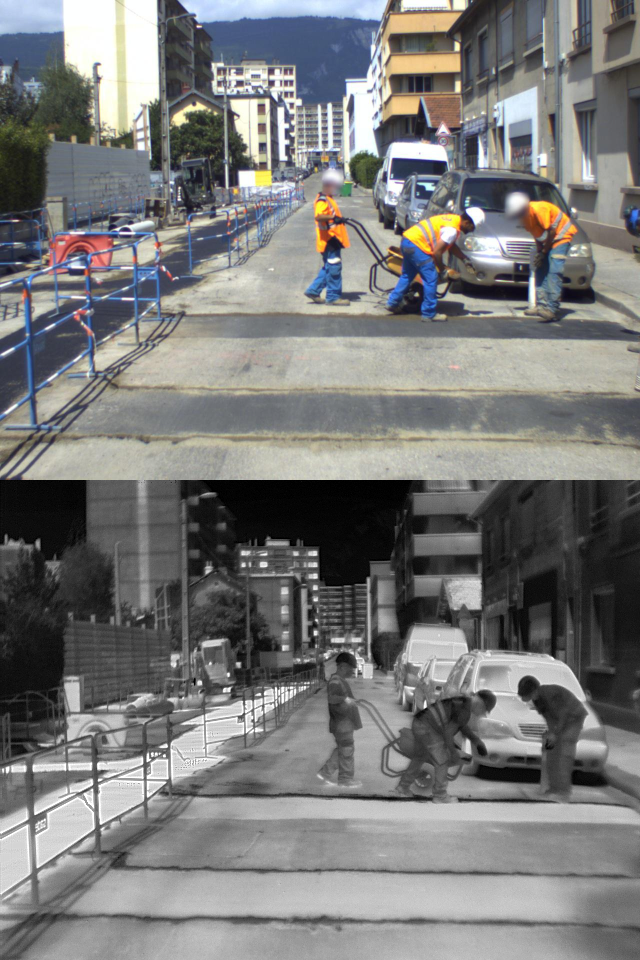}
        \caption{External temperature: \num{30}\si{\celsius}}
    \end{subfigure}
    \hfill
    \begin{subfigure}[t]{0.32\linewidth}
        \centering
        \includegraphics[width=\linewidth]{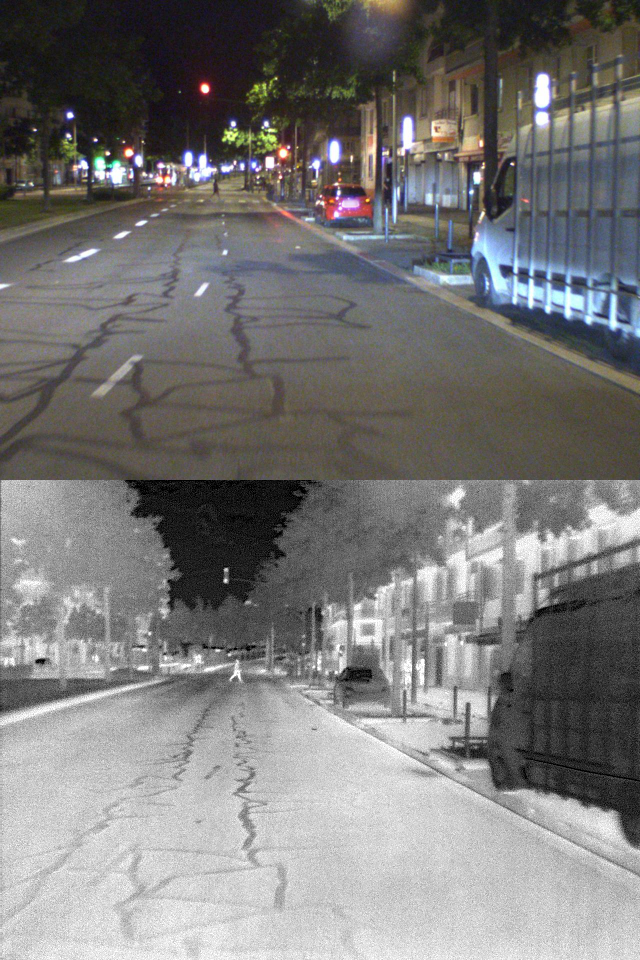}
        \caption{External temperature: \num{20.6}\si{\celsius}}
    \end{subfigure}
    \caption{Images from a broad range of seasonal and heat scenarios are present in the LYNRED mobility dataset, offering challenging conditions for developing infrared-based AEB.}
    \label{fig:temperaturecontrasts}
\end{figure*}

Images were acquired all year round, covering a wide range of ambient temperatures from 3.7\si{\celsius} to 30\si{\celsius}. This temperature variability presents a significant challenge for thermal or long-wave infrared (LWIR) sensors since they operate by capturing thermal radiation. When environmental temperatures approach that of the human body, the contrast between objects and their surroundings decreases, potentially affecting detection performance ; see Figure \ref{fig:temperaturecontrasts} for an example.\\

\begin{figure*}[!ht]
    \centering
    \begin{subfigure}{1\linewidth}
        \centering
        \includegraphics[trim=0 1cm 0 1cm, width=0.99\linewidth]{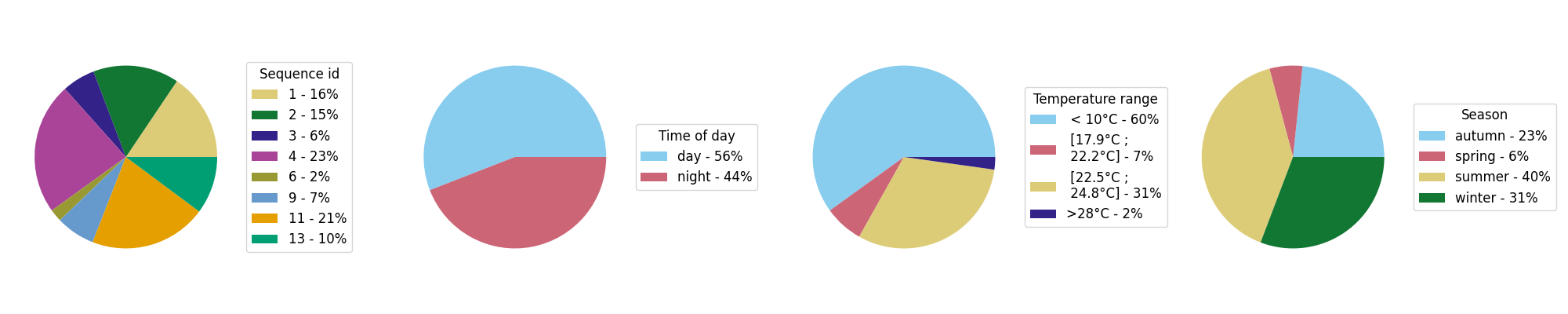}
        \caption{Train dataset}
    \end{subfigure}
    \begin{subfigure}{0.99\linewidth}
        \centering
        \includegraphics[trim=0 1cm 0 .5cm, width=1\linewidth]{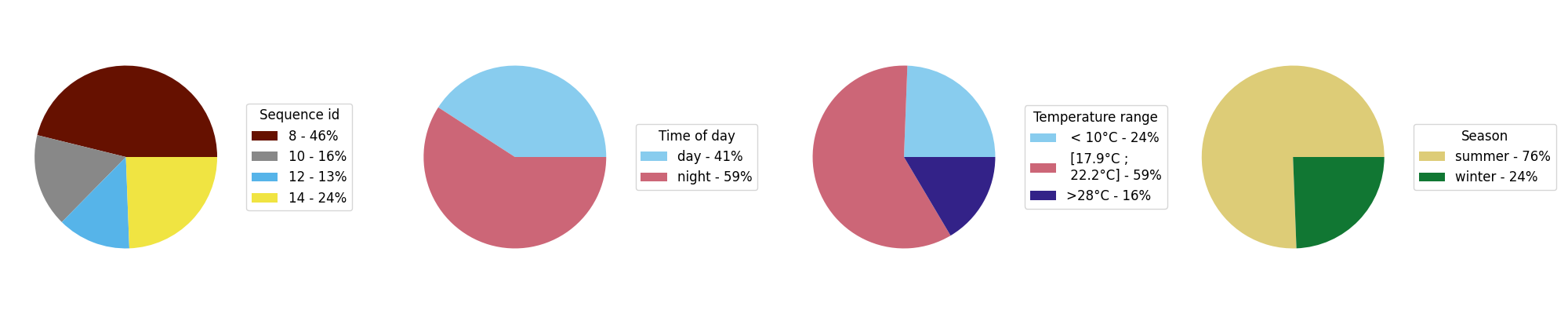}
        \caption{Test dataset}
    \end{subfigure}
    \caption{Decomposition of the different data splits.}
    \label{fig:train-test-distrib}

\end{figure*}

To prevent data leakage, the sequences were split into separate training and testing subsets while ensuring a representative distribution of time of day and seasonal conditions. The distribution of the dataset split across various parameters is illustrated in Figure~\ref{fig:train-test-distrib}.\\

Additionally, 9 different classes are annotated to reflect the diversity of road users. A comprehensive comparison of class labels across the existing datasets' training sets is presented in Table~\ref{tab:classimbalance}. Fewer classes were chosen to counteract class imbalance. Indeed, 6 out of the 16 classes of FLIR have under 100 samples in the train set and most of these are not present in the test set~\cite{FLIR}. Furthermore, the occlusion status of human-class objects is also recorded, following the method used in other existing infrared object detection datasets~\cite{FLIR, jia2021llvip, chen2024amfd}. The full annotation protocol is described in the section 'Validation and Quality'.  \\

\begin{table*}[!h]
    \centering
    \resizebox{\textwidth}{!}{
        \begin{tabular}{cccccccccccccccc}
        \hline
        \multirow{2}{*}{Datasets} & \multicolumn{15}{c}{Number of instances per class}    \\ 
         &  person & bicycle &  car &  motorcycle &  bus &   train &   truck &   light &   hydrant &   sign &   Animal  &   skateboard &    stroller &    scooter &  other vehicle \\ \hline

           KAIST \cite{hwang2015multispectral} &   103128 &   - &   - &  - &   - &   - &   - &   - &   - &   - &   - &    - &   - &   -   &  - \\

                FLIR ADAS \cite{FLIR} &   50478 &  7237  & 73623  & 1116 & 2245 & 5 &  829 & 16198 & 1095  &  20770 & 12 $\star$ & 29  &  15  &  15    & 1373  \\

                LLVIP \cite{jia2021llvip} &  34137  &   - &   - &   - &   - &   - &   - &   - &   - &   - &   - &   - &   - &    -   &  -  \\

                M3FD \cite{m3fd} &  7443  &   - &  11737 &  344   &  449  &  - &  654  & 1542  &   - &   - &   - &   - &   - &   -   &  -  \\

                KAIST sanitized \cite{chen2024amfd} & 24247  &   - &   - &   - &   - &   - &   - &   - &   - &   - &   - &   - &   - &   -   &  -  \\

                FLIR aligned \cite{zhang2020multispectral} & 8987 & 2566 & 20608 &   - &   - &   - &   - &   - &   - &   - &    - &   - &   - &   -   &  -  \\

                LYNRED-MDS (ours) &  9152 & 1755 & 15924 &  310  &  244	  &   304 &   514 &   - &   - &   - &  80 & - &   - &   -   & 91   \\ \hline
        \end{tabular}}
    \caption{Number of instances per class across the different reference  RGB-T datasets. The number of annotations displayed corresponds to the infrared train split. We notice that even though FLIR offers far more classes than others, the class imbalance makes most of those useless for training. The Starred $\star$ value represents blended classes (4 dogs and 8 deers).}
    \label{tab:classimbalance}
\end{table*}

We provide both 16-bit-per-pixel and processed 8-bit-per-pixel infrared images, as tone mapping has been shown to be a relevant component of the embedded object detection pipeline~\cite{karam2024optimizing}. The 16-bit images correspond to the minimally processed output of the infrared camera.\\

\begin{figure*}[!htbp]
   \begin{subfigure}{0.32\linewidth}
        \centering
        \includegraphics[width=1 \linewidth]{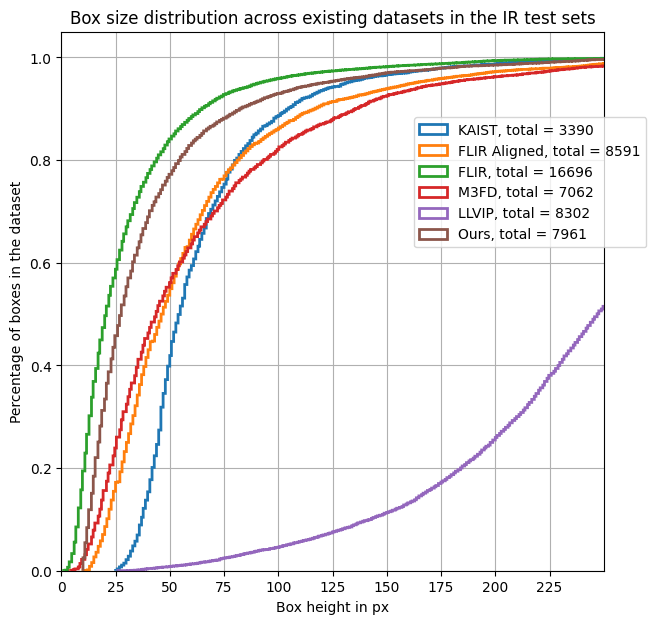}
        \caption{}
        \label{fig:Cumulative_all}
    \end{subfigure}
    \hfill
    \begin{subfigure}{0.32\linewidth}
        \centering
        \includegraphics[width=1\linewidth]{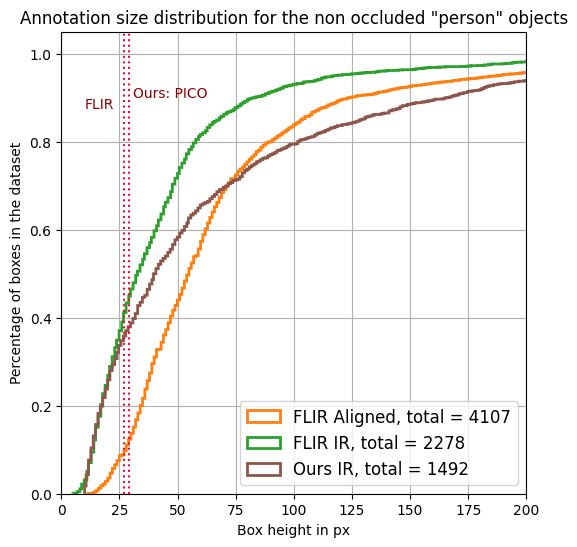}
        \caption{}
        \label{fig:Cumulative_FLIRvs}
    \end{subfigure}
    \hfill
    \begin{subfigure}{0.32\linewidth}
        \centering
        \includegraphics[width=0.95\linewidth]{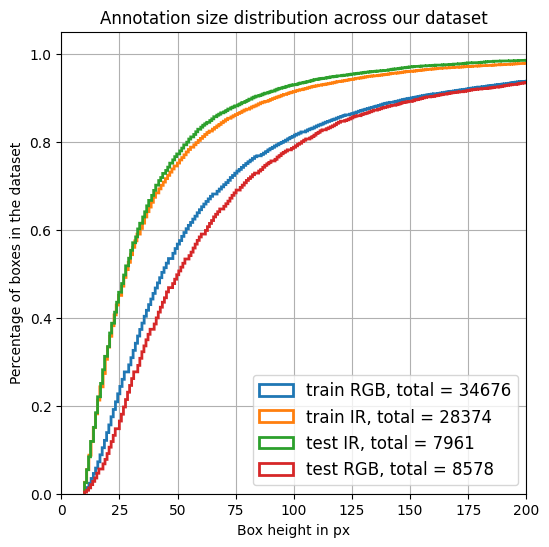}
        \caption{}
    \end{subfigure}
    \caption{Cumulative histograms of bounding boxes height for all presented datasets (a), comparison between FLIR ADAS, FLIR Aligned and LYNRED-MDS IR splits (b)  and LYNRED-MDS's RGB and IR modalities (c). The gap between IR and RGB annotation sizes for curve (c) can be explained by the difference between the IR and RGB sensors specifications.}
    \label{fig:Cumulative}
\end{figure*}

In Figure~\ref{fig:Cumulative}, we plot the empirical cumulative distribution of bounding box height for each dataset; this allows us to pinpoint some important differences between the datasets concerning object size annotations. We can clearly see that our dataset is on par with most other RGB-T datasets, resembling M3FD the most with respect to its bounding box height distribution. The task shift between LLVIP and the others is easily noticeable, as objects in video surveillance clips are at a controlled distance from the camera, leading to a smaller variance and overall bigger boxes.\\

Finally, in the multimodal detection subset of LYNRED-MDS objects as small as 10 pixels in height are annotated. This is done to evaluate the efficiency of infrared sensors for pedestrian detection in the context of regulatory standards cited in the introduction~\cite{NHTSA}. To assess the capability of detecting pedestrians at a distance of 50 meters, we estimate the expected pixel height $h_{\text{object}}$ of an unobstructed human in our dataset by $h_{\text{object}} = \frac{f \times H_{\text{object}} \times h_{\text{image}}}{D_{\text{object}}\times H_{\text{sensor}}}$ derived from~\cite{szeliski2022computer} and where $H_x$ and $h_x$ refer to physical height (mm) or height in pixels, respectively.\\

For an average adult pedestrian height provided by NHTSA as $H_{\text{object}} = 1.750\si{\meter}$~\cite{NHTSAheight} and in order to meet regulatory requirements with our sensor setup specifications, we compute a necessity to detect objects with a minimum height of approximately 28.8 pixels on the PICO sensor (focal length $f = 14\si{\milli\meter}$, image height $h_{\text{image}} = 480\text{ pixels}$, sensor height $H_{\text{sensor}} = 8.16\si{\milli\meter}$, object distance $d_{\text{object}} = 50\si{\meter}$), respectively 40.8 pixels for the ATTO sensor (sensor height $H_{\text{sensor}} = 5.76\,\text{mm}$).\\

Applying the same calculation to FLIR’s dataset yields a comparable value of 26.8 pixels. We also performed the computation for a child--sized pedestrian (approximately 100\si{\centi\meter} tall), which establishes more challenging detection threshold of 16.5 and 23.3 pixels for the PICO and the ATTO sensors, respectively. Meanwhile the value would be 11.9 pixels in FLIR's dataset.\\

In Figure~\ref{fig:Cumulative_FLIRvs}, we illustrate the pixel height thresholds for all unobstructed pedestrian annotations in the FLIR ADAS validation set, as well as the FLIR Aligned and LYNRED-MDS test datasets. Notably, in both datasets, more than 35\% of annotated humans are 29 pixels tall or smaller. Per NHTSA regulations \cite{NHTSA}, a vehicle traveling at 100km/h must detect pedestrians at a minimum of 50 meters to allow sufficient braking time, making this the most demanding AEB scenario — successful detections at 50m implying successful detections at shorter distances and lower speeds. These datasets thus provide a challenging benchmark for AEB-relevant small pedestrian detection, especially in critical situations where detecting small objects remains a significant challenge for object detection algorithms \cite{gundougan2023ir, li2021yolo, wei2024review}, though a comprehensive AEB evaluation would additionally require consideration of motion patterns and spatiotemporal continuity. \\

Images were anonymized using the blurring algorithm from~\cite{anonymizer} targeting licence plates and faces to make the dataset compliant with the regulations of the CNIL (\emph{Commission Nationale de l’Informatique et des Libertés}), the French data protection authority.

\section*{VALIDATION AND QUALITY} 

\subsection*{Annotation protocol and guidelines}

Annotations were performed by a professional labelling company and subsequently verified by an in-house
expert in infrared imaging. Annotators labeled RGB and infrared images separately and independently,
annotating objects exclusively in the modality where they are visible. Bounding boxes were required to be
as tight as possible, with a minimum object height of 10 pixels. Every annotation carries a mandatory
occlusion tag: None (fully visible), Partial (more than 30\% visible), or Heavy (less than 30\% visible). Objects were only annotated if identifiable by a human annotator. Nine object classes were defined: person,
car (smaller than a Mercedes Sprinter), truck (bigger than a Mercedes Sprinter), bus, bicycle, motorcycle, train (including tramways), animal, and construction machine, with a
special tag used for atypical instances. Riders were annotated separately from their vehicles and tagged
accordingly. Bicycles and motorcycles, were annotated as such even if not ridden. 

The dataset repository contains two annotation versions \texttt{metadata} and \texttt{metadata\_small\_objects}. The \texttt{metadata\_small\_objects} 
folder contains the original annotations from a first expert phase, which include 
objects as small as 4 pixels in height. However, since this finer labelling was only 
applied to a subset of the data, it introduces inconsistencies that can lead to 
unreliable evaluation and are only kept as legacy. The \texttt{metadata} folder contains the unified  annotation set, in which all bounding boxes below 10 pixels in height are excluded 
regardless of their phase of origin, ensuring full consistency with the protocol 
described above. All experiments and benchmarks reported in this study are based 
exclusively on the \texttt{metadata} annotations.

\subsection*{Experimental results : Unimodal cross-dataset evaluation}

\begin{table*}[!h]
  \centering
  \begin{subtable}{0.49\linewidth}
        \centering
        \resizebox{\linewidth}{!}{
        \begin{tabular}{|c|cccccc|}
            \cline{2-7}
            \multicolumn{1}{l|}{} & \multicolumn{6}{c|}{Train dataset} \\
            \hline
              Test dataset & ALL & LLVIP & FLIR & FLIR Aligned & M3FD & \textbf{ours}\\
            \hline
              ALL & \cellcolor{gray!60}  \textbf{0.88} & 0.61 & 0.54 & 0.54 & 0.59 & \cellcolor{gray!30}  0.73 \\

              LLVIP & \cellcolor{gray!60}  \textbf{0.96} & \cellcolor{gray!60}  \textbf{0.96} & 0.41 & 0.44 & 0.64 & \cellcolor{gray!30}  0.77 \\
                
              FLIR ADAS & 0.61 & 0.11 & \cellcolor{gray!60}  \textbf{0.80} & 0.55  & 0.43 & \cellcolor{gray!30}  0.62 \\
              
              FLIR Aligned & \cellcolor{gray!60}  \textbf{0.85} & 0.22 & 0.79 & \cellcolor{gray!30}  0.81 & 0.48 & 0.76 \\
              
              M3FD & \cellcolor{gray!30}  0.85 & 0.29 & 0.67 & 0.54 & \cellcolor{gray!60} \textbf{ \textbf{0.86}} & 0.61 \\
              
              \textbf{ours} & \cellcolor{gray!60}  \textbf{0.65} & 0.18 & 0.49 & 0.45 & 0.34 & \cellcolor{gray!30}  0.60 \\
             \hline 
        \end{tabular}
        }
        \caption{IR}
        \label{tab:map50_humanIR}
    \end{subtable}
    \hfill
    \begin{subtable}{0.49\linewidth}
        \centering
        \resizebox{\linewidth}{!}{
         \begin{tabular}{|c|cccccc|}
            \cline{2-7}
            \multicolumn{1}{l|}{} & \multicolumn{6}{c|}{Train dataset} \\
            \hline
            Test dataset & ALL & LLVIP & FLIR & FLIR Aligned & M3FD & \textbf{ours} \\
            \hline
            ALL & \cellcolor{gray!60}  \textbf{0.77} & 0.54 & 0.36 & 0.31 & \cellcolor{gray!30}  0.55 & 0.45 \\
            LLVIP & \cellcolor{gray!60}  \textbf{0.97} & \cellcolor{gray!30}  0.54 & 0.17 & 0.11 & 0.36 & 0.34 \\
            
            FLIR ADAS & 0.44 & 0.14 & \cellcolor{gray!60}  \textbf{0.70} & 0.39 & 0.39 & \cellcolor{gray!30}  0.52 \\
            
            FLIR Aligned & \cellcolor{gray!60}  \textbf{0.69} & 0.24 & 0.58 &\cellcolor{gray!30}  0.64 & 0.51 & 0.47 \\
            
            M3FD & \cellcolor{gray!30}  0.74 & 0.23 & 0.42 & 0.24 & \cellcolor{gray!60} \textbf{ 0.77} & 0.38 \\
            
            \textbf{ours} & \cellcolor{gray!30}  0.49 & 0.07 & 0.36 & 0.19 & 0.22 & \cellcolor{gray!60} \textbf{0.51} \\
            \hline
        \end{tabular}
        }
        \caption{RGB}
        \label{tab:map50_humanRGB}
    \end{subtable}
  \caption{Mean Average Precision (mAP@50) for human detection across~\subref{tab:map50_humanIR} IR and~\subref{tab:map50_humanRGB} RGB datasets. Only the 'person' class is retained for comparison purposes. Each column uses a same training dataset and each row uses a same evaluation dataset. The "ALL" dataset combines LLVIP, FLIR Aligned, M3FD, and ours. All experiments were conducted using YOLOv8n trained for 200 epochs. For each row, the best result is highlighted in bold and dark gray; the second best is highlighted in light gray.}
  \label{tab:map50_human}
\end{table*}

To benchmark our dataset, we evaluated the finetuning performances of YOLOv8n~\cite{Jocher_Ultralytics_YOLO_2023}, pretrained on the COCO dataset~\cite{lin2014microsoft}. We selected this model for its ease of use and suitability for deployment. The model is fine-tuned for 200 epochs with a batch size of 16, using the "auto" optimizer setting provided by Ultralytics. \\

Rather than focusing solely on in-domain performance, our work emphasizes the model's generalization capabilities. To assess this, we adopted mAP@50 (mean Average Precision at an Intersection-over-Union threshold of 0.5) as our evaluation metric. This threshold is commonly used in object detection benchmarks and offers a reasonable balance between detection quality and tolerance for localization error, particularly important when detecting small or low-contrasted objects in thermal imagery.\\

Our generalization test protocol involves training the model on one dataset and evaluating it on the others. We focus exclusively on the "person" or "human" class, as it is the most relevant for VRU detection tasks. The datasets used in our experiments are FLIR ADAS, FLIR Aligned, M3FD, LLVIP, and ours. Additionally, we include a combined dataset, referred to as ALL, which merges LLVIP, FLIR Aligned, M3FD, and ours. This serves as a topline reference. The comparative results across datasets are presented in Table~\ref{tab:map50_human}.\\

It highlights several key findings. First, the model trained on the "ALL" dataset achieves the best performance on its constituent datasets : LLVIP, FLIR Aligned, M3FD, and ours. In contrast, performance on FLIR ADAS -- excluded from the "ALL" set -- is notably lower. This, as well as FLIR Aligned's performance on FLIR ADAS's test set, suggests that FLIR Aligned may not fully capture the data distribution of the broader FLIR ADAS dataset.\\

Second, we observe a modality gap: IR generally outperforms RGB across datasets in object detection. This trend is especially pronounced in datasets such as LLVIP, which focuses on nighttime or low-light conditions where RGB sensors are less effective. Interestingly, despite LLVIP's size being far larger than most other datasets, its models perform poorly in cross-dataset validation. This emphasizes the relevance of task-specific datasets focusing on driving scenarios.\\

These preliminary results, suggest that LYNRED-MDS offers competitive generalization potential among the compared datasets. The model trained on our IR images achieves the second-best cross-dataset mAP on three out of five target datasets, and the LYNRED-MDS test split consistently ranks among the more challenging evaluation sets for models trained on other driving datasets. While these observations are encouraging, broader conclusions about generalization would require validation across additional architectures and metrics. Nevertheless, LYNRED-MDS appears to be a promising and challenging benchmark for infrared-based pedestrian detection in ADAS contexts.

\section*{RECORDS AND STORAGE} 

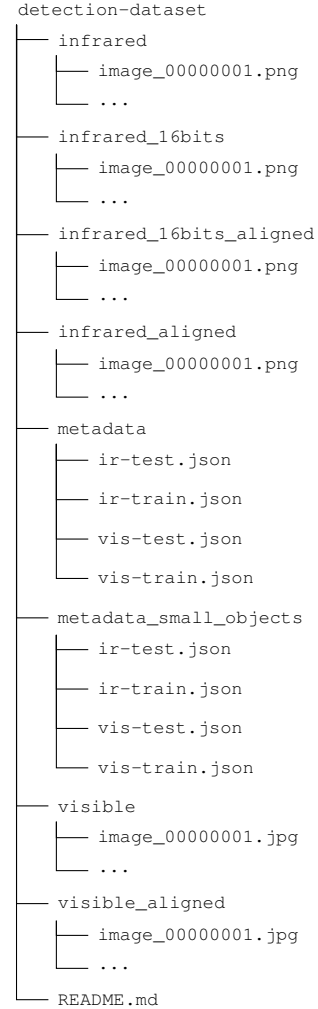
\begin{figure}
    \centering
    \begin{forest}
        for tree={
            font=\scriptsize\ttfamily,
            grow'=0,
            child anchor=west,
            parent anchor=south,
            anchor=west,
            calign=first,
            s sep=2pt,
            l=10pt,
            edge path={
                \noexpand\path [draw, \forestoption{edge}]
                (!u.south west) ++(3pt,0) |- (.child anchor) \forestoption{edge label};
            },
            before typesetting nodes={
                if n=1
                    {insert before={[,phantom]}}
                    {}
            },
            fit=band,
            before computing xy={l=15pt},
        }
        [detection-dataset
          [infrared
            [image\_00000001.png]
            [\dots]
          ]
          [infrared\_16bits
            [image\_00000001.png]
            [\dots]
          ]
          [infrared\_16bits\_aligned
            [image\_00000001.png]
            [\dots]
          ]
          [infrared\_aligned
            [image\_00000001.png]
            [\dots]
          ]
          [metadata
            [ir-test.json]
            [ir-train.json]
            [vis-test.json]
            [vis-train.json]
          ]
          [metadata\_small\_objects
            [ir-test.json]
            [ir-train.json]
            [vis-test.json]
            [vis-train.json]
          ]
          [visible
            [image\_00000001.jpg]
            [\dots]
          ]
          [visible\_aligned
            [image\_00000001.jpg]
            [\dots]
          ]
          [README.md]
        ]
    \end{forest}
    \caption{Overview of the structure of the dataset folder.}
    \label{fig:dataste_view}
\end{figure}

Data is stored on LYNRED's website (\url{https://www.lynred.com/lynred-mobility-dataset}), alongside 2 other datasets. The dataset is organised in 6 folders as described in Figure~\ref{fig:dataste_view}. The folders tagged '\_aligned'  correspond to images that have been geometrically aligned through the chessboard method described earlier. The dataset annotation data is presented using the standard COCO JSON format \cite{lin2014microsoft} to which we add a few additional fields. These fields are described in Table~\ref{tab:annot-fields}.

\begin{table}[h]
    \centering
        \begin{tabularx}{\linewidth}{lX}
        \hline
        \textbf{Field} & \textbf{Description / Use} \\
        \hline
        \multicolumn{2}{l}{\textbf{Images}} \\
        \hline
        id & Unique identifier of the image \\
        width, height & Image resolution in pixels \\
        file\_name & File name of the thermal image (PNG) \\
        visible\_image & Corresponding RGB image file (JPG) \\
        season & Season of acquisition (e.g., summer, winter) \\
        time\_of\_day & Time of acquisition (e.g., day, night) \\
        tamb & Ambient temperature at acquisition (°C) \\
        sequence\_id & Sequence identifier (groups images from same sequence) \\
        author & Source/creator of the data (LYNRED or Neovision) \\
        \hline
        \multicolumn{2}{l}{\textbf{Annotations}} \\
        \hline
        id & Unique identifier of the annotation \\
        image\_id & Reference to the corresponding image (via \texttt{images.id}) \\
        category\_id & Object class (links to \texttt{categories.id}) \\
        bbox & Bounding box coordinates [x, y, width, height] in pixels \\
        area & Area covered by the bounding box \\
        iscrowd & COCO convention flag (0: normal object, 1: crowd/ambiguous) \\
        \hline
        \end{tabularx}
    \caption{Description of JSON fields for \texttt{images} and \texttt{annotations}.}
    \label{tab:annot-fields}
\end{table}

\section*{INSIGHTS AND NOTES} 

\subsection*{Multispectral object detection}
Although not used in this study, the paired images can be used for multispectral object detection. In that sense, it is comparable to other commonly used datasets such as FLIR Aligned and M3FD. To further facilitate the use of the dataset in multispectral object detection, it is planned to release an aligned version that corrects parallax for close objects.

\subsection*{Tone mapping}
The release of 16-bit `RAW' infrared images allows their use for tone-mapping-free object detection, further reducing computational load in an AEB scenario, as well as enabling the development of optimal tone mapping algorithms for object detection tasks.

\subsection*{Regional adaptation}

While our dataset reflects Western European driving conditions and vehicle standards, we acknowledge that adaptation to other regional contexts — such as different vehicle typologies, road infrastructure, or climate profiles — may require additional data collection or domain adaptation techniques. We encourage the community to build upon this dataset and extend it to other geographical contexts using the protocol provided in this paper.

\section*{SOURCE CODE AND SCRIPTS} 

The code used to obtain Figure \ref{fig:train-test-distrib}, \ref{fig:Cumulative}, and Table \ref{tab:map50_human} can be found on this github repository : \url{https://github.com/arbezlo/LYNRED_Mobility_Dataset} .

\section*{ACKNOWLEDGEMENTS}

\text{\hspace{1em}} L.A. was the main author of the manuscript and was responsible for the data split and the experiments. J.M. and X.B. handled dataset collection and supervised labelling. All authors contributed to the manuscript.
\newline

\text{\hspace{1em}} The article authors have declared no conflicts of interest.
\\
\bibliographystyle{IEEEtran}
\bibliography{references_corrected}

\end{document}